\documentclass[conference]{IEEEtran}
\IEEEoverridecommandlockouts
\usepackage[utf8]{inputenc}
\usepackage{caption}
\usepackage{subcaption}
\usepackage{graphicx}
\usepackage{amsmath}
\usepackage{multirow}
\usepackage[export]{adjustbox}
\usepackage{cite}
\usepackage{amsmath,amssymb,amsfonts}
\usepackage{algorithmic}
\usepackage{graphicx}
\usepackage{textcomp}
\usepackage{xcolor}
\usepackage[hidelinks]{hyperref}
\def\BibTeX{{\rm B\kern-.05em{\sc i\kern-.025em b}\kern-.08em
    T\kern-.1667em\lower.7ex\hbox{E}\kern-.125emX}}

\makeatletter
\newcommand{\linebreakand}{%
  \end{@IEEEauthorhalign}
  \hfill\mbox{}\par
  \mbox{}\hfill\begin{@IEEEauthorhalign}
}
\makeatother

\begin{document}

\title{Predicting Personas Using Mechanic Frequencies and Game State Traces}
% \author{\IEEEauthorblockN{Anonymous Authors}}
\author{
\IEEEauthorblockN{Michael Cerny Green}
\IEEEauthorblockA{
\textit{Game Innovation Lab} \\
\textit{New York University, Tandon}\\
New York City, USA \\
mike.green@nyu.edu}
\and
\IEEEauthorblockN{Ahmed Khalifa}
\IEEEauthorblockA{
\textit{Institute of Digital Games} \\
\textit{University of Malta}\\
Msida, Malta \\
ahmed@akhalifa.com}
\and
\IEEEauthorblockN{M Charity}
\IEEEauthorblockA{
\textit{Game Innovation Lab} \\
\textit{New York University, Tandon}\\
New York City, USA \\
mlc761@nyu.edu}
\linebreakand
\IEEEauthorblockN{Debosmita Bhaumik}
\IEEEauthorblockA{
\textit{Game Innovation Lab} \\
\textit{New York University, Tandon}\\
New York City, USA \\
debosmita.bhaumik01@gmail.com}
\and
\IEEEauthorblockN{Julian Togelius}
\IEEEauthorblockA{
\textit{Game Innovation Lab} \\
\textit{New York University, Tandon}\\
New York City, USA \\
julian@togelius.com}
}

\maketitle

% discuss where playstyles come from: either from designer theories (e.g. Bartle's typology, newer stuff) or unsupervised learning (e.g. SOMs in Tomb Raider)
% we are here exploring defining playstyle as deviation from what a specialized synthetic agent would do (AAR)
% and seeing if we can predict this in various ways from simpler metrics (mechanic activation frequency)
%there is a need for fast, accurate prediction
% we also see if we can predict self-labeling and if self-labeling correlates with agent-defined playstyle

%mike: run unsupervised learning (clustering) on mechanic activation frequencies on the user data
%mike: get statistics on different multi-label combinations of users

\begin{abstract}
We investigate how to efficiently predict play personas based on playtraces. Play personas can be computed by calculating the action agreement ratio between a player and a generative model of playing behavior, a so-called procedural persona. But this is computationally expensive and assumes that appropriate procedural personas are readily available. We present two methods for estimating play personas, one using regular supervised learning and aggregate measures of game mechanics initiated, and another based on sequence learning on a trace of closely cropped gameplay observations. While both of these methods achieve high accuracy when predicting play personas defined by agreement with procedural personas, they utterly fail to predict play style as defined by the players themselves using a questionnaire. This interesting result highlights the value of using computational methods in defining play personas.

%\todo[inline]{New abstract using pieces of old abstract}
%In this paper, we promote the idea that game mechanic frequencies are good at capturing the essence of human playstyles.
%Human playtraces are labeled by calculating a user's action agreement with artificial agent personas, which we use as a ground truth. First, we build and run artificial A* agents on 5 maps within the Minidungeons 2 framework to build a synthetic dataset. We perform a user study to collect human playtrace data on the 5 maps, while also capturing a self-perceived label of their own playstyle using a questionnaire. We then train a classifier on a user's game mechanic frequency vector and an action-agreement multi-label.
%We demonstrate that a classifier trained on synthetic data alone fails to generalize out of distribution on the human-based test set. However, a classifier trained on human data is able to perform well on the human-based test set, demonstrating that mechanic frequency vectors are a good representation of playstyle, that global behavior of synthetic agents is not representative of human behavior while local behavior of agents is a good approximation. In a curious finding, we discover that participants are not perceptive of their playstyle, often believing they play a certain way while behaving differently. We demonstrate this by contrasting discrepancies between self-labels and action agreement labels.

\end{abstract}


\section{Introduction}

% Mike: the following is a paraphrase of the procedural personas background section, i really like how it dives into where personas come from, we can edit (or remove if we dont like it). Its probably too detailed for our purposes, which also needs to talk about mechanics and player classification as personas USING personas
% The use of computational methods to imbue computer game characters with personality has been a focus of game AI programming since the very beginning of the medium.
% As one instance, Short provides an overview of how non-player characters can be provided with human-like personalities under the heading of procedural personalties \cite{short2016procedural}.

The use of personas has a long history within design in general and design for information technology in particular. The approach was pioneered for software development in the early 1990s \cite{cooper2004inmates} as a method for structuring and operationalizing qualitative data gathered from design research, chiefly in the form of interviews.
% Based off interview data, a number of personas would be defined.
% Each of these would serve as a specific instantiation of groups of user concerns that tended to co-occur, expressed as an archetypal example user, fully fleshed out with names, back stories, concerns and preferences.
Canossa and Drachen transported this approach into the realm of game design \cite{canossa2009patterns}, defining personas less in terms of general life concerns and more in terms of player interaction preferences within the space of the game. They call their conceptualization \emph{play personas} and operationalized their definition through data mining, suggesting how the persona design process could be supported by analyzing quantitative game data gathered via telemetrics \cite{tychsen2008defining}.
While play personas are archetypal models of player behavior inferred from experience or observed data, the re-projection into the game itself is something that is done imaginatively by the designer(s) of the game: i.e. play personas let us understand what players \emph{have done}, but do not enact what players \emph{might do}.
\emph{Procedural personas}~\cite{holmgard2015evolving,holmgard2014generativeagents,holmgaard2014personas} extend the play persona idea by adding a game-playing, generative aspect.
By capturing persona characteristics from designer specification or from observed data, and formalizing these as utility functions, procedural personas are implemented as agents that can act in the game, enabling automatic playtesting.

% Experience Driven procedural content generation~\cite{shaker2010towards, yannakakis2011experience} seeks to generate game content conditional on preferences, personas, or abilities of particular players or player archetypes. Applications range from the automated generation of levels targeting specific personas \cite{liapis2015procedural} and mechanics \cite{khalifa2019intentional,green2018generating,green2020mario} to tutorials targeting new players~\cite{green2018atdelfi}. Being able to rapidly and accurately classify a player's play persona would be a valuable tool for personalized content generative systems.

In this paper, the use of personas comes full circle. Hand-designed utility functions steer artificial agents to behave in a specific way. These agents are then used to classify human playtraces with a ground truth label using Action Agreement Ratios (AAR)~\cite{holmgaard2014personas}. Though accurate, AAR calculations can be expensive for long playtraces. Agents must calculate their moves at every gametick. For tree-search and evolutionary agents, the allotted search time is typically proportional to their performance: the longer time given to search, the better they perform. Often times in the industry, game studios have hundreds to thousands of trace telemetry datapoints streaming in by the hour. For example, King, the developer of Candy Crush Saga,
%\footnote{https://www.king.com/game/candycrush},
reports a daily average of 13 million active players\footnote{https://activeplayer.io/candy-crush-saga/}. For large or streaming datasets, calculating AAR can be an incredibly expensive computational challenge, even with parallelization. We propose training a machine learning algorithm on known playtraces to rapidly classify unseen playtraces with an acceptable level of accuracy when compared to AAR. Classifying new playtraces does not require the use of agent play, so it has the potential to be much faster than AAR. 

Additionally, we propose to use a mechanic frequency vector as an aggregated representation for the playtraces instead of game states. A mechanic frequency vector is an integer vector that represents the number of times each mechanic is activated during a playtrace. When compared to game state traces, a mechanic frequency vector is usually a smaller representation of a playtrace. To demonstrate the speed and accuracy of mechanic frequency vectors as a representation, we train classifiers on human playtrace data using AAR labels, and infer on a testing set of more human playtraces using AAR labels. For comparison, we also train a classifier on the actual playtrace (sequence of game states) and compare its results to the ones trained on the mechanic frequency vector. Additionally, we train both classifiers on synthetic generated data using the same personas and on the human playtrace data labeled by humans themselves. While the classifiers trained on synthetic data and the classifiers trained on self-labeled human data fail to generalize to the test set, the classifiers trained on AAR labeled human data is able to accurately classify human playtraces.

\section{Background}
The following subsections review game mechanic definitions and analysis, artificial persona research, and finally a brief description of the Minidungeons 2 game, within which this research was performed.

\subsection{Game Mechanic Analysis}
Though there is no universally accepted definition for what a ``game mechanic'' is, there are several popular definitions. Jarvinen~\cite{jarvinen2008games} characterizes game mechanics as functional game features which allow the player to interact with the game state for achieving a desired goal state. Sicart~\cite{Sicart2008DefiningGM} defines game mechanics as an action which is triggered by an agent to interact with the game world. 
% \todo[inline]{MC: Is this sentence needed? The subgroups of mechanics are never mentioned again aside from this paragraph. If we keep it, can we at least label the MiniDungeons mechanics into one of these 3 groups or say how one mechanic might be primary for a persona but secondary for another?}
% Additionally, he defines three groups with which to classify mechanics: core mechanics are game mechanics which are used repeatedly by agents to reach a "systemically rewarded end-state"; primary mechanics are core mechanics that agents can use directly to solve challenges, which lead to the desired end state; secondary mechanics are core mechanics which help the agent to reach the end-state. Salen and Zimmerman~\cite{salen2003rules} describe core mechanics as "the essential play activity" which the player performs over and over in the game. 
% Each of these definitions are environmentally centric: they define mechanics in terms of environmental manipulation. To add another dimension, Green et al.~\cite{green2021game} proposes Game Mechanic Alignment Theory as a framework through which to organize mechanics through the lens of systemic rewards and agential motivation to examine why players want to trigger mechanics. 

%Talk about all the various ways to analyze game mechanics: critical mechanic discovery methods (uninformed - Press space, Atdelfi; informed - critical mechanic discovery from playtraces), building levels from mechanics (mario levels, mech elites), 

There are many ways to use mechanics to analyze player behavior and even generate content. Green et al.~\cite{green2017press,green2018atdelfi} define ``critical mechanics'' as the set of game mechanics which need to be triggered to reach the win state. They use search based methods to find the critical mechanics to use as an input for automated video game tutorial generation. Silva et al. generate compact heuristics (i.e. mechanical instructions) for blackjack~\cite{silva2016blackjack} and post-flop poker~\cite{silva2018poker}.
Khalifa et al.~\cite{khalifa2019intentional} generate Super Mario Bros (Nintendo, 1985) level snippets called ``scenes,'' which spotlight the use of mechanics to complete successfully, which Green et al.~\cite{green2020mario} build upon to develop entire levels in the Mario AI Framework~\cite{togelius2010mario} using ``scene stitching.'' Charity et al.~\cite{charity2020mech} uses the quality diversity algorithm known as ``Map Elites'' to illuminate the mechanic design space for several video games within the GVGAI framework. Mechanic Miner~\cite{cook2013mechanic} evolves mechanics for 2D puzzle-platform games, using Reflection to find a new game mechanic then generate levels that utilize it. With the the \emph{Gemini} system~\cite{summerville2017mechanics}, users can input mechanics to generate information about the game's affordances~\cite{mateas2001preliminary}. Mappy~\cite{osborn2017automatic} and the updated Mappyland~\cite{osborn2021mappyland} can transform a series of button presses into a graph of room associations, transforming movement mechanics into level maps for Nintendo Entertainment System games.

None of the above research proposes using mechanical frequencies as an aggregated representation of an entire playtrace, which is what we do in this paper. However, they do use mechanics to perform complex analysis on players, to provide content for players, and to estimate player behavioral patterns, which supports our proposed representation.

\subsection{AI Personas}
Video games allow players to experience levels designed by the games' creators in almost an infinite number of ways. Each player has a distinct style, method, and strategy for navigating through a gamespace which we will refer to as a persona. Bartle~\cite{bartle1996hearts} originally proposed a taxonomy of players and their personas based on how they interacted with the environment and other players ranging from killers, socializers, achievers, and explorers. Many game designers try to design their levels to cater to a specific persona or to design levels with multiple paths that work with different personas. %As such,%some designers perform behavior analysis on game levels to generate data-driven personas such as
Various attempts have been made to infer play personas through unsupervised learning from playtraces, including early work using data from
Tomb Raider: Underworld~\cite{drachen2009player} and Starcraft~II~\cite{avontuur2013player}.

Tychsen and Canossa~\cite{tychsen2008defining} first introduced the concept of defining play personas through game metrics and mechanics performed. These metrics could be used to recreate a specific player but may not give as much insight as to what a type of player might do on a general scale (i.e. how would the same player react in a game level they have never played before.) With automated and artificial personas, these agents can be inserted into the game to predict how a player may approach a game level, saving human and computation resources and speeding up the design process. Procedural agents developed via evolution~\cite{holmgaard2014evolving,holmgaard2018automated} and reinforcement learning~\cite{holmgard2014generative} have shown to be accurate in emulating player behaviors in a game setting. These artificial personas can be used for automated playtesting and encapsulate a variety of behaviors which lead to a wider diversity of levels designed for players with different play personas.

Action Agreement Ratio (AAR) is a metric of tracking behavioral similarity between playtraces proposed by Holmg{aa}rd et al~\cite{holmgaard2014personas,holmgaard2015monte}. Generating AAR between a human and an agent is done by first reconstructing every game state of the human playtrace. At each state, the persona/agent being tested receives the state as input and calculates its next move. If the human action and the agent action are in agreement at that state, the agent's agreement metric is increased by $1$. After running over every state, the AAR is calculated by dividing the total score by the number of moves made. We use AAR as a ground truth in this paper to label human-based data.

\subsection{MiniDungeons 2}
MiniDungeons 2~\footnote{http://minidungeons.com/} is a 2-dimensional deterministic, turn-based, rogue-like game, first described in \cite{holmgaard2015minidungeons}, in which the player takes on the role of a hero traversing a dungeon level, with the end goal of reaching the exit. Typically set on a 10 by 20 tile grid, the game map is made up of a mix of impassible \textit{walls} and passable \textit{floors}. Interactive items and characters are scattered throughout the level. To win, the player must reach the exit, represented as a staircase. All game characters have Hit Points (HP) and deal damage when the player collides with them. The player begins the game with 10 HP, and if the player runs out of HP, they die and lose the level.

Each turn, the player selects an action to perform, and all game characters will then move after the player completes their action. Any game character may move in one of the four cardinal directions (North, South, East, West) on their turn as long as the tile in that direction is not a wall. The player is given one re-usable javelin at the start of every level. The player may choose to throw this javelin and do 1 damage to any monster within their unbroken line of sight. After using the javelin, the hero must traverse to the tile to which it was thrown in order to pick it up and use it again.

% game objects
While exploring a map, the player can find interactive objects that result in various effects:
\begin{itemize}
\item\textbf{Potions} increase the HP of the hero by 1, up to the max of 10. 
\item\textbf{Treasures} increase the treasure score.  
\item\textbf{Portals} come in pairs. Heroes can use to transport themselves to the paired portal on the same turn.
\item\textbf{Traps} deal 1 damage to any game character moving through them, including monsters and enemies.
\end{itemize}

% game characters
In addition to the above objects, the player may encounter monsters, all of which desire to attack a player within line of sight and some of which have additional secondary goals:
\begin{itemize}
\item\textbf{Goblins} move 1 tile every turn towards the player if within line of sight. They have 1 HP and deal 1 damage upon collision. Goblins try to avoid colliding with other goblins and goblin wizards.
\item\textbf{Goblin Wizards} cast a 1 damage spell at the hero if they have line of sight within 5 tiles of the player. If they are over 5 tiles from the player but have line of sight, they will move 1 tile towards the player each turn. Wizards have 1 HP and deal no damage on collision.
\item\textbf{Blobs} move toward either a potion or the hero within line of sight. They will move 1 tile towards the closest one per turn, preferring potions over the hero in case of a tie. A blob colliding with a potion or another blob consumes it and enables it to level-up into a more powerful blob. The default, lowest level blob has 1 HP and does 1 damage upon collision. The 2\textsuperscript{nd} level blob has 2 HP and does 2 damage. The most powerful blob has 3 HP and does 3 damage. 
\item\textbf{Ogres} will move 1 tile towards either the player or a treasure per turn within line of sight, preferring treasures over the hero in case of a tie.  When an ogre collides with a treasure, they consume it, and their sprite becomes fancier to look at. Ogres have 2 HP and deal 2 damage to anything they collide with, including other ogres.
\item\textbf{Minitaurs} always move 1 step along the shortest path to the hero as determined by A* search, regardless of line of sight. Collision with the minitaur will deal 1 damage to the player. A minitaur has no HP and is immortal. When damaged, the minitaur is stunned for 5 moves.
\end{itemize}

\section{MiniDungeons 2 Playtraces}\label{sec:playtraces}
In a Minidungeons 2 gameplay session, players usually move around with their character to interact with different game objects and characters. At every frame, we record the current game state as a 2D map of tiles, the player's current health, score, location, and last performed action. We also record all the interactions (game mechanics) that happens in the game when the player interacts with game objects and characters. Within the scope of this paper, there are a total of 17 mechanics tracked during play:
\begin{itemize}
    \item \textbf{Enemy Kill:} the player slays any enemy
    \item \textbf{Monster Hit:} a specific monster is hit by either the player or the javelin. This mechanic is recorded separately for each different monster so we have \textbf{Goblin Hit}, \textbf{Minitaur Hit}, \textbf{Goblin Wizard Hit}, \textbf{Blob Hit}, or \textbf{Ogre Hit}.
    \item \textbf{Ogre Treasure:} the ogre collects a treasure
    \item \textbf{Blob Potion:} the blob consumes a potion
    \item \textbf{Blob Combine:} the blob combines with another blob
    \item \textbf{Javelin Throw:} the player throws the javelin
    \item \textbf{Collect Treasure:} the player collects a treasure
    \item \textbf{Consume Potion:} the player consumes a potion
    \item \textbf{Trigger Trap:} a trap is triggered
    \item \textbf{Use Portal:} the player uses a portal
    \item \textbf{End Turn:} the player makes a move and ends their turn
    \item \textbf{Die:} the player dies
    \item \textbf{Reach Stairs:} the player reaches the exit stairs
\end{itemize}

\subsection{Synthetic Dataset Generation}\label{sec:synthetic-traces}
We curate a group of artificial agent personas to generate the synthetic playtraces. These agents are Best First Search (BestFS) agents which use utility functions created by Holmg{aa}rd et al~\cite{holmgaard2014personas,holmgaard2018automated} in order to resemble a player archetype (persona). Previous research~\cite{holmgaard2018automated} uses MCTS agents, but in a fully deterministic game, BestFS generates more consistent results, both for action agreement and for gameplay ability. There are three agent personas in MiniDungeons that are used to capture agent playtraces with the following described goals:
% \todo[inline]{J: Should we mention something about computation time? Also that these are novel, because the previous ones used MCTS, but in a fully deterministic game A* turned out to be bette?R}

\begin{enumerate}
    \item \textbf{Runner:} complete the level as fast as possible.
    \item \textbf{Monster Killer:} slay as many monsters as possible before completing the level.
    \item \textbf{Treasure Collector:} open as many chests as possible before completing the level.
\end{enumerate}
Agents perform online planning to move, meaning that each turn, they build a search tree to find the optimal next move to make. After doing preliminary experiments, we realized that the synthetic agents \textit{always} make the same movements, and therefore multiple playtraces on the same map will not expand the dataset. To overcome this, we added random movement to the synthetic agents when generating the synthetic dataset: every move there is a 25\% chance that the agent will take a random action rather than their searched action. This randomness is \textit{not} used when performing action agreement. 

At each turn, the BestFS agent uses the following heuristic for each different persona. In the following equations, $h_{agent}$ denotes the heuristic function for that specific BestFS agent. The agents could be runner ($r$), monster killer ($mk$), or treasure collector ($tc$). The runner agent is simply getting to the exit in the fewest amount of steps as possible. Equation~\ref{eq:r} shows the heuristic ($h_{r}$) function of the runner agent.
\begin{equation}\label{eq:r}
    \begin{aligned}
   h_{r} = dist_{exit} - steps
    \end{aligned}
\end{equation}
where  $dist_{exit}$ is the distance from the current player location to the exit, and $steps$ is the amount of steps taken since the start of the game.

MK agents will prioritize killing all monsters while attempting not to die and getting to the exit when all monsters have been slain. Equation~\ref{eq:mk} shows the heuristic ($h_{mk}$) function of the monster killer agent.
\begin{equation}\label{eq:mk}
    \begin{aligned}
   h_{mk} = c * N_{monster} + k * Dead + p_{monster}\\
   p_{monster} = \begin{cases} 
     min(dist_{monster}) & N_{monster} > 0 \\
     dist_{exit} & N_{monster} == 0
      \end{cases}
    \end{aligned}
\end{equation}
where $c$ and $k$ are constants, $min(dist_{monster})$ is the distance to the closest monster from the player location, $dist_{exit}$ is the distance between the player and the exit, $N_{monster}$ is the number of alive monsters in the level, and $Dead$ is a binary value that is equal to 1 if the player is dead and 0 otherwise.

TC agents will prioritize collecting treasure while attempting not to die and getting to the exit when all treasures have been collected. It uses a similar equation to~\ref{eq:mk} but replacing $min(dist_{monster})$ with $min(dist_{treasure})$ (the distance to the closest treasure from the player location) and $N_{monster}$ with $N_{treasure}$ (number of unopened treasures in the level).

% Equation~\ref{eq:tc} shows the heuristic ($h_{tc}$) and the cost ($g_{tc}$) functions of the treasure collector agent.
% \begin{equation}\label{eq:tc}
%     \begin{aligned}
%   h_{tc} = \begin{cases} 
%      min(dist_{treasure}) & N_{treasure} > 0 \\
%      dist_{exit} & N_{treasure} == 0
%   \end{cases}\\
%   g_{tc} =  c * N_{treasure} + k * Dead
%     \end{aligned}
% \end{equation}
% where c and k are constants, $min(dist_{treasure})$ is the distance to the closest treasure from the player location, $dist_{exit}$ is the distance between the player and the exit, $N_{treasure}$ is the number of unopened treasures in the level, and $Dead$ is a binary value that is equal to 1 if the player is dead and 0 otherwise.

\subsection{User Study}\label{sec:human-traces}
We prepare the following user study to gather human playtraces and to capture how users perceive their own play personas. The study is split into two parts, the first of these being a questionnaire and the second a gameplay session to gather playtraces. The questionnaire part consists of $10$ multiple choice questions. 
For the questionnaire section, the first question of the study (``How often do you play games?'') is to get insight into the player's familiarity with gaming.
Answer choices for this question are the following: Never, Once or twice a month, Once a week, A few times a week, Every day. This question is not specific about the genre or type of game being played, which can vary greatly.

The following list of questions contained within the gameplay questionnaire section of the user study are pertinent to the user's self-perception of their play persona:
\begin{enumerate}
    \setcounter{enumi}{1}
    \item I try to finish the level (or a run of the game) as fast as possible
    \item I look around a game area for all the hidden treasures or items before proceeding
    \item I try to defeat/kill all the enemies I can see
    \item I enjoy battling enemies
    \item I enjoy searching for hidden treasures or items
    \item I only care about finishing the level
    \item I make sure to collect every collectable object I can
    \item I try to move through a level as move-efficient as possible
    \item I like fighting different kinds of enemies and collecting experience
\end{enumerate}
Answer choices for the above questions are the following:  Never, Rarely, Sometimes, Often, and Always. There are three questions which pertain to battling enemies, three questions about collecting treasures, and three questions about speed running. The purpose of these questions is to gain insight into the player's perception of their own gameplay goals for dungeon-crawler, rogue-like games such as Minidungeons 2. Questions $2$, $7$, and $9$ are inquiring about ``runners'', questions $3$, $6$, and $8$ about ``treasure collectors'', and questions $4$, $5$, and $10$ about ``monster killers''. 

After finishing the questionnaire, the player starts playing the game. There are $3$ tutorial levels to acclimate the user to the game. These levels are very simple, beginning with a simple small map and slowly increasing in size and mechanical complexity until introducing all the basic mechanics by the 3rd level. The user is then presented with a series of levels randomly selected from the maps described in Section \ref{sec:experiment-maps}.

\section{MiniDungeon 2 Labels}\label{sec:labels}
The collected playtraces in MiniDungeon 2 are labeled using different methods depending on the source of the data. Synthetic data is generated from the procedural personas, therefore their labels are already known. On the other hand, user data can be labeled either from the results coming from the questionnaire - the self-perceived label - or by comparing the user's actions with the procedural agents' actions - the action agreement label. While the synthetic data will always have a singular persona label, however the user data may have multiple play personas and thus can be labeled as multiple personas or possibly none at all.

\subsection{Self Perceived Label}
The self-perceived user playtraces labels are calculated from the answers the user selected for the questionnaire given at the beginning of the user study. These answers are transformed to numbers where $Always$ is equal to $4$ and $Never$ is equal to $0$. We average these across their related questions:
questions 2, 7, and 9 are averaged under the "Runner" score; questions 3, 6, and 8 are averaged under the "Treasure Collector" score; and, questions 4, 5, and 10 are averaged under the "Monster Killer" score. At first, we decided to convert this average to a label by checking if the average value is greater than 2 (over 50\%), in which case the associated persona is added to the user's self-perception multilabel. We found that users tend to rank themselves high for most questions (section~\ref{sec:res_ques}) causing more than 70\% of all playtraces to be labeled Runner, Treasure Collector, and Monster Killer all at the same time. 
As a consequence, we decide to compare each average value to the average rating over all the user questionnaires as a form of normalization. If a participant's score in a given persona category is greater than the average rating for that category over all questionnaires, the persona label is added to the participant. After performing this calculation, every playtrace of the participant is assigned the participant's labels.

\subsection{Action Agreement Label}
The artificial persona agents described in Section \ref{sec:synthetic-traces} are used to generate action agreement labels for individual playtraces in both the human training set and the human testing set. This is done using a relatively simple replay process:
Given a playtrace, a replay agent steps through every move that the human made. At every step, the system runs each of the 3 generative agents for $1$ second, searching for the moves they would have made at that step. This means that each move takes $3$ seconds to calculate agreement. User playtraces can be as long as $140$ moves or more, meaning that a single playtrace could take several minutes to compute without parallelization. If the user's action agrees with an agent's action, that persona gets a point. The persona's label is added to the playtrace if they agree more than they disagree with that persona, i.e. if they have more than 50\% action agreement calculated by taking that persona's point score divided by total steps taken.

\section{Experiment}\label{sec:experiment}
In this section, we describe our experiment setup in details. We begin with the Minidungeons 2 maps that were selected and why. We then define our agent setup and finally explain the classifiers that are trained on synthetic and human playtraces.

\subsection{Minidungeons 2 Maps}\label{sec:experiment-maps}
\begin{figure*}[ht]
    \centering
    \begin{subfigure}[b]{0.18\textwidth}
   \centering
   \includegraphics[height=25ex]{images/maps/Map100.png}
   \caption{Map 100}
   \label{fig:map100}
    \end{subfigure}
    \begin{subfigure}[b]{0.18\textwidth}
   \centering
   \includegraphics[height=25ex]{images/maps/Map101.png}
   \caption{Map 101}
   \label{fig:map101}
    \end{subfigure}
    \begin{subfigure}[b]{0.18\textwidth}
   \centering
   \includegraphics[height=25ex]{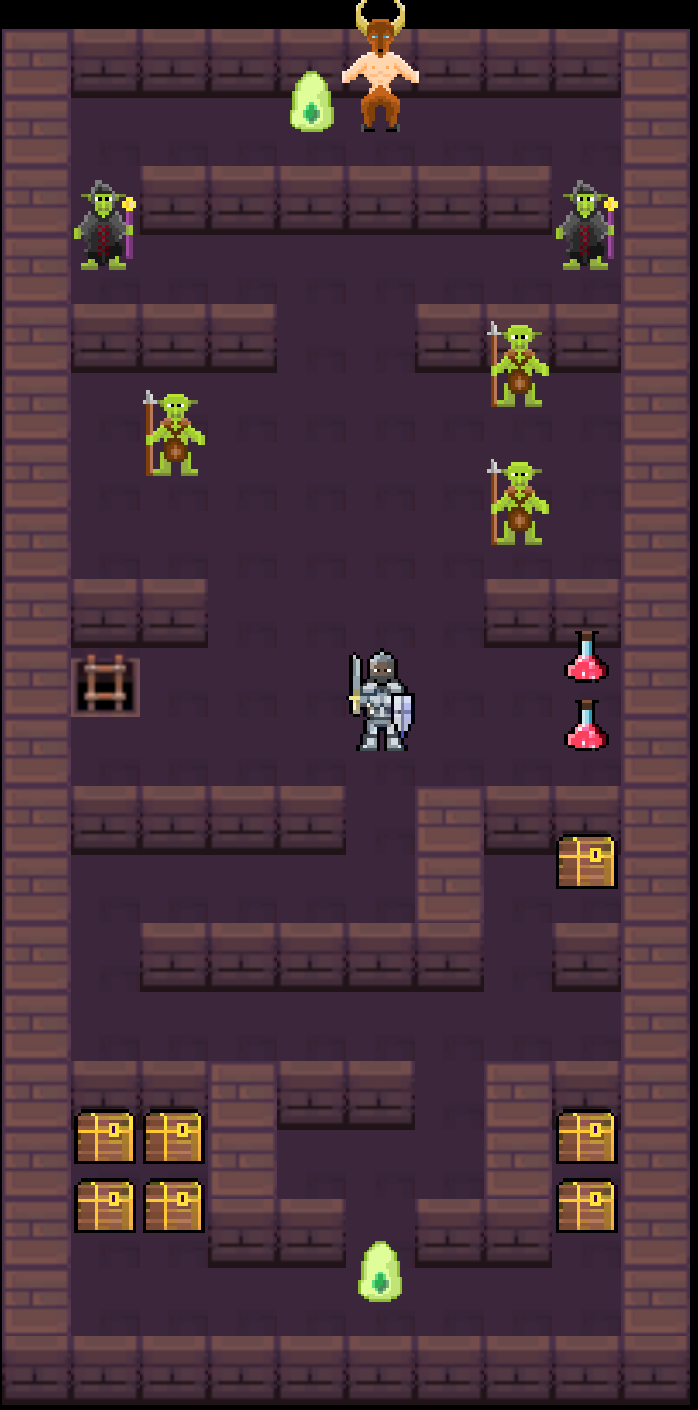}
   \caption{Map 102}
   \label{fig:map102}
    \end{subfigure}
    \begin{subfigure}[b]{0.18\textwidth}
   \centering
   \includegraphics[height=25ex]{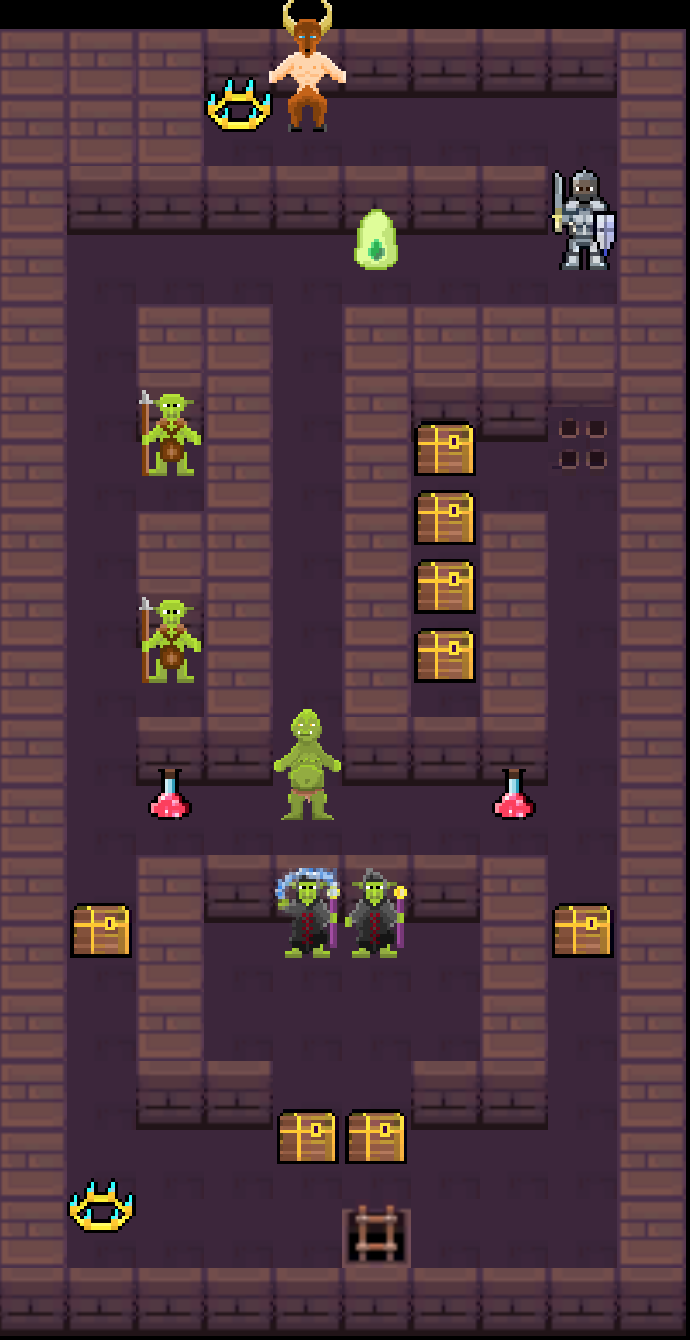}
   \caption{Map 201}
   \label{fig:map201}
    \end{subfigure}
    \begin{subfigure}[b]{0.18\textwidth}
   \centering
   \includegraphics[height=25ex]{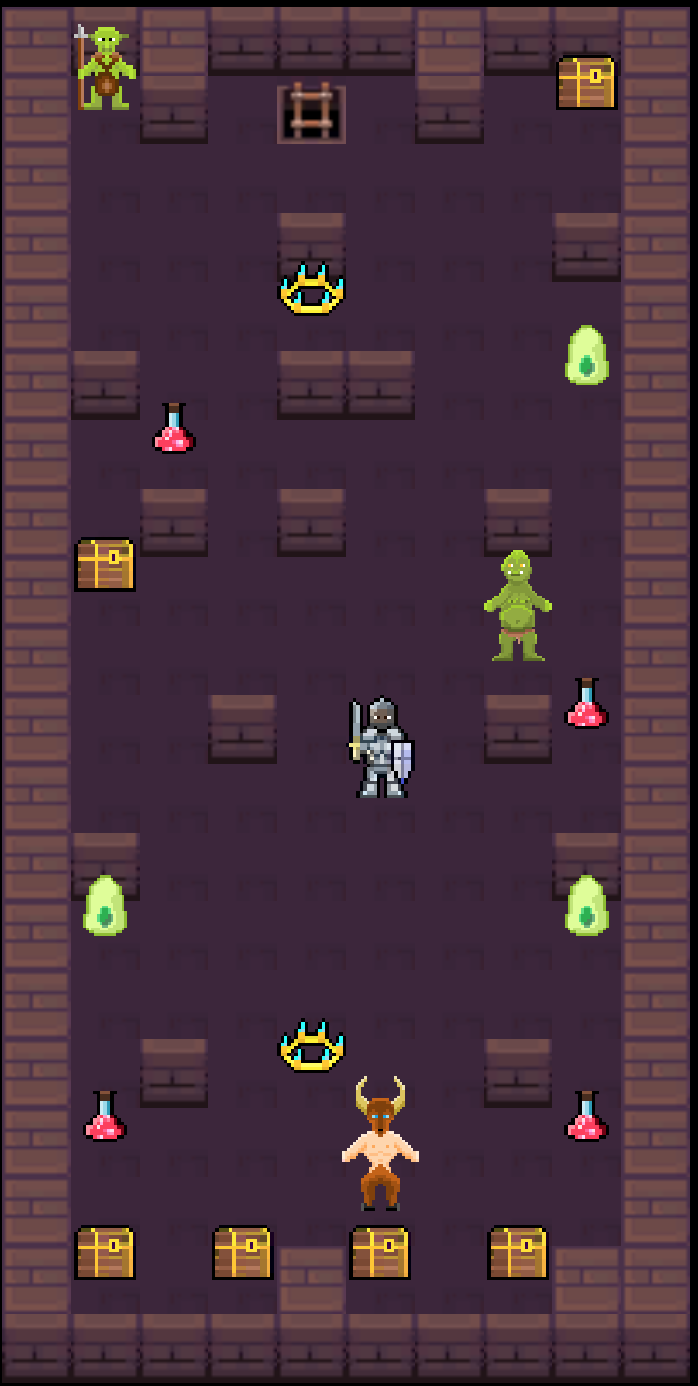}
   \caption{Map 202}
   \label{fig:map202}
    \end{subfigure}
    \caption{User study maps used to generate playtrace datasets}
    \label{fig:game_maps}
    \vspace{-15pt}
\end{figure*}

Five hand-designed maps were developed for this project. All maps contained a mix of game elements and offered multiple routes for the player to take to complete the level. Figure~\ref{fig:game_maps} shows all the five levels, as we can see all the maps except Map 202 provide very clear three paths for each different persona. For example, Map 100 has a center path for runner, right path for treasure collector, and left path for monster killer. Meanwhile, Map 202 is an open space with monsters and treasures distributed all over the map. Every map contains $5-6$ monsters, $6-9$ treasures, and at least one straightforward path to the exit.

\subsection{Agent Setup}
All agents are given $1.0$ second to plan before making their next move. For the monster killer agent and treasure collector, we used $c$ equal to $45$ and $k$ equal to $double.max$ to encourage the agent to stay alive and favor states that kill more monsters in case of monster killer or collect more treasure in case of treasure collector. Each persona runs 100 times each on the five map described in Section \ref{sec:experiment-maps}, for a total of 1500 playtraces.

\subsection{User Study Data}
A link to the user study described in Section \ref{sec:human-traces} was spread by a combination of tweets and academic email lists. $205$ participants completed the whole study, while an additional $158$ only played the game and did not answer the questionnaire. We were able to extract $565$ playtraces to use as training set data among the participants. Our test data is made of additional $293$ playtraces from participants, which brings our total dataset size to $853$.

\subsection{Classifying Players}
Players can be classified as Runners (R), Treasure Collectors (TC), or Monster Killers (MK) using a machine learning model trained on labeled playtraces. We trained multiple different models and tested them on a common test set consisting of $293$ playtraces that are labeled using AAR. The training dataset composed of $565$ playtraces is divided into a 70-30 split for training and validation for all of the models.

We train our models on two different types of inputs:
\begin{itemize}
    \item \textbf{Cropped playtraces:} We convert the states of the playtraces into a smaller state by cropping the states to focus on the 3x3 area around the player. 
    This particular size was used because player interactions and the mechanics examined for the MiniDungeons game only affect the immediate surrounding area of the player (i.e. collecting nearby treasure or potions or attacking.) The smaller size also prevents the machine learning model from overfitting and ideally helps it generalize better~\cite{ye2020rotation}.
    % reworded the following -MC
    % The reason behind the conversion is the player interactions/mechanics only affect the surrounding area and doesn't affect anything else on the map. Also, using cropped states helps machine learning models to generalize instead of overfitting on the training data
    \item \textbf{Mechanic frequencies:} We use the frequency of the 17 triggered mechanics specified in Section \ref{sec:playtraces} as the input for our model. By ``frequency'' we mean the exact number of times each mechanic is triggered over the course of one level. These mechanics are represented in vector format, and normalized on a mechanic-basis across the entire dataset before training.
\end{itemize}

These two input representations are used for all 3 datasets described in sections~\ref{sec:playtraces} and \ref{sec:labels}:
\begin{enumerate}
    \item The synthetic agent dataset with labeled personas.
    \item The user dataset with self-perceived labels calculated from the questionnaire.
    \item The same user dataset as above but labeled by Action Agreement measured using the previously mentioned synthetic personas. 
    % \todo[inline]{MC: Needs distinguishing. Is it the synthetic agent personas (from \#1) or the replay agents?}
\end{enumerate}

Based on these different input data representation, we decided to train two different types of models:
\begin{itemize}
    \item \textbf{Long Short Term Memory (LSTM) Model:} this model is trained on the cropped playtraces. LSTM can train and generalized on sequential data. The model consists of a LSTM layer with 100 hidden nodes followed by an output layer of 3 nodes for each persona probability. The LSTM model is trained using stochastic gradient descent for 200 epochs and learning rate of 0.001. For each experiment, we train 3 networks to make sure the results are stable.
    \item \textbf{Support Vector Machine (SVM) Model:} this model is trained on the mechanic frequencies. We picked SVMs because they are a well established machine learning method that is known for its capability to generalize on small datasets without overfitting.
    % \todo[inline]{MC: Citation?}
\end{itemize}

\section{Results and Discussion}

In this section, we review the results of the user study questionnaire detailed in Section \ref{sec:human-traces} and discuss the results of the classifier training explained in Section \ref{sec:experiment}. 

\subsection{User Study Questionnaire}\label{sec:res_ques}
% \begin{figure}[ht]
%     \centering
%     \includegraphics[width=\columnwidth]{images/questionaire/gaming_breakdown.png}
%     % \todo[inline]{J: this figure takes too much space, which we could use for other things; a table would be smaller.}
%     \caption{``How often do you play video games?''}
%     \label{fig:gaming_breakdown}
% \end{figure}

The responses to the question ``How often do you play video games'' demonstrate that over $89.3\%$ ($184$) of participants play video games once a week or more. We can assume from this that the vast majority of participants are at least familiar with optimizing gameplay actions for the purpose of achieving some goal, such as winning.
% \todo[inline]{Any more demographic data on the players?}
% \todo[inline]{Mike: no, :(}

\begin{figure}[ht]
    \centering
    \includegraphics[width=0.85\columnwidth,valign=t]{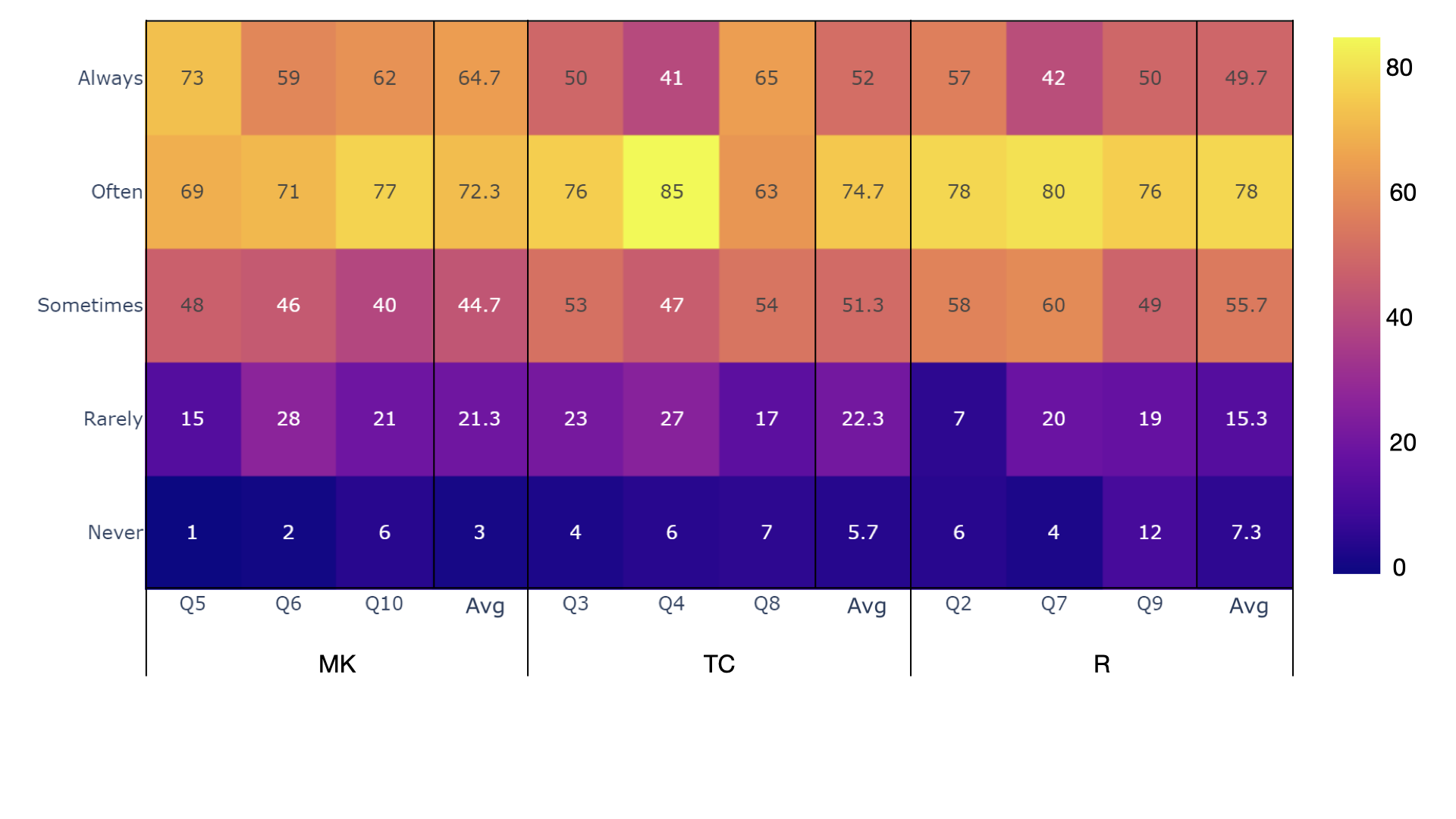}
    \vspace{-15pt}
    \caption{Questionnaire breakdown totals. From left to right, columns are grouped as MK, TC, and R questions.}
    %A majority of answers are \textit{Sometimes}, \textit{Often}, or \textit{Always}.}
    \label{fig:questionaire_breakdown}
\end{figure}

Figure \ref{fig:questionaire_breakdown} shows the results for all these questions and their averages. The numbers/color in each cell corresponds to the number of users with that response. We can see that most users answer Always, Often and Sometimes, suggesting that they see themselves as a mix of different personas rather than one in particular. Out of the $206$ participants, $154$ perceived themselves to be Runners, $151$ perceived themselves to be Treasure Collectors, and $159$ perceived themselves to be Monster Killers, according to their self-perception multilabel.

\subsection{AAR and Self-Perceived Label Analysis}\label{sec:label-comparison}

\begin{table*}[ht]
\centering
\begin{subtable}[ht]{0.49\textwidth}
    \resizebox{\textwidth}{!}{%
        \begin{tabular}{|r|l|l|l|l|}
        \hline
        \multicolumn{1}{|l|}{} & \textit{Count} & \multicolumn{1}{c|}{\textit{Steps Taken}} & \multicolumn{1}{c|}{\textit{Treasures Collected}} & \multicolumn{1}{c|}{\textit{Enemies Killed}} \\ \hline\hline
        \textit{No Label} & 82    & 21.46$\pm$30.77    & 0.6$\pm$1.55     & 1.21$\pm$2.29    \\ \hline\hline
        \textit{Pure R}   & 144   & \textbf{18.49$\pm$8.07}     & 0.48$\pm$1.19    & 1.47$\pm$1.69    \\ \hline
        \textit{Pure TC}  & 64    & 73.84$\pm$29.98    & \textbf{5.75$\pm$1.79}    & 3.48$\pm$2.16    \\ \hline
        \textit{Pure MK}  & 2     & 92.5      & 5.0     & \textbf{2.5}     \\ \hline\hline
        \textit{R \& TC}  & 76    & \textbf{36.82$\pm$23.67}    & \textbf{3.36$\pm$3.1}     & 2.91$\pm$1.96    \\ \hline
        \textit{R \& MK}  & 56     & \textbf{15.25$\pm$9.61}      & 0.18$\pm$0.95     & \textbf{1.45$\pm$1.32}     \\ \hline
        \textit{TC \& MK} & 24    & 113.17$\pm$19.03   & \textbf{6.83$\pm$0.56}    & \textbf{5.0$\pm$2.3}      \\ \hline\hline
        \textit{R \& TC \& MK} & 117   & \textbf{43.95$\pm$23.59}    & \textbf{3.63$\pm$2.86}    & \textbf{3.85$\pm$1.96}    \\ \hline
        \end{tabular}%
    }
    \caption{Action Agreement Labels}
    \label{tab:aar-label}
\end{subtable}
\begin{subtable}[ht]{0.49\textwidth}
    \resizebox{\textwidth}{!}{%
        \begin{tabular}{|r|l|l|l|l|}
        \hline
        \multicolumn{1}{|l|}{} & \textit{Count} & \multicolumn{1}{c|}{\textit{Steps Taken}} & \multicolumn{1}{c|}{\textit{Treasures Collected}} & \multicolumn{1}{c|}{\textit{Enemies Killed}} \\ \hline\hline
        \textit{No Label} & 176    & 28.35 $\pm$ 26.27    & 1.44 $\pm$ 2.38     & 1.98$\pm$1.89    \\ \hline\hline
        \textit{Pure R}   & 42   & \textbf{17.95$\pm$8.75}     & 0.36$\pm$0.57    & 1.47$\pm$1.63    \\ \hline
        \textit{Pure TC}  & 49    & 39 $\pm$ 33.94    & \textbf{2.62 $\pm$ 3.11}    & 2.43 $\pm$ 2.27    \\ \hline
        \textit{Pure MK}  & 54     & 34.17 $\pm$ 34.31      & 1.78 $\pm$ 2.23     & \textbf{2.22 $\pm$ 2.41}     \\ \hline\hline
        \textit{R \& TC}  & 11    & \textbf{21.71 $\pm$ 23.03}    & \textbf{1.03 $\pm$ 2.15}     & 0.84 $\pm$ 0.99    \\ \hline
        \textit{R \& MK}  & 17     & \textbf{37.92 $\pm$ 30.05}      & 2.61 $\pm$ 2.7     & \textbf{3.8 $\pm$ 2.13}     \\ \hline
        \textit{TC \& MK} & 49    & 53.96 $\pm$ 41.69   & \textbf{3.85$\pm$3.21}    & \textbf{2.82 $\pm$ 2.18}      \\ \hline\hline
        \textit{R \& TC \& MK} & 169   & \textbf{40.86$\pm$30.69}    & \textbf{3.18$\pm$3.14}    & \textbf{3$\pm$2.41}    \\ \hline
        \end{tabular}%
    }
    \caption{Self Perceived Labels}
    \label{tab:self-label}
\end{subtable}
\caption{Average scores and standard deviations in several game metrics for every AAR and Self Perceived multilabel combination on the $565$ human playtrace training dataset. The positive correlation mechanics for each label are \textbf{bold}.}
% AAR labeled Pure MK has no standard deviation due to low count.}
\label{tab:aar-sl-label-breakdown}
\end{table*}

\begin{table*}[ht]
\centering
\begin{subtable}[ht]{0.25\textwidth}
    \centering
    \begin{tabular}{|r|lcc|}
    \hline
    \multicolumn{1}{|l|}{}        & \multicolumn{3}{c|}{\textit{Self-Percieved}}                \\ \hline
    \multirow{3}{*}{\textit{AAR}} & \multicolumn{1}{l|}{}    & \multicolumn{1}{c|}{Yes}  & No   \\ \cline{2-4} 
                                  & \multicolumn{1}{c|}{Yes} & \multicolumn{1}{c|}{29.6\%} & 40.0\% \\ \cline{2-4} 
                                  & \multicolumn{1}{c|}{No}  & \multicolumn{1}{c|}{12.7\%} & 17.7\% \\ \hline
    \end{tabular}
    \label{tab:r-agreement}
    \caption{Runner}
\end{subtable}
\begin{subtable}[ht]{0.25\textwidth}
    \centering
    \begin{tabular}{|r|lcc|}
    \hline
    \multicolumn{1}{|c|}{}        & \multicolumn{3}{c|}{\textit{Self-Percieved}}                \\ \hline
    \multirow{3}{*}{\textit{AAR}} & \multicolumn{1}{l|}{}    & \multicolumn{1}{c|}{Yes}  & No   \\ \cline{2-4} 
                                  & \multicolumn{1}{c|}{Yes} & \multicolumn{1}{c|}{29.0\%} & 20.7\% \\ \cline{2-4} 
                                  & \multicolumn{1}{c|}{No}  & \multicolumn{1}{c|}{20.2\%} & 30.1\% \\ \hline
    \end{tabular}
    \label{tab:tc-agreement}
    \caption{Treasure Collector}
\end{subtable}
\begin{subtable}[ht]{0.25\textwidth}
    \centering
    \begin{tabular}{|r|lcc|}
    \hline
    \multicolumn{1}{|l|}{}        & \multicolumn{3}{c|}{\textit{Self-Percieved}}                \\ \hline
    \multirow{3}{*}{\textit{AAR}} & \multicolumn{1}{l|}{}    & \multicolumn{1}{c|}{Yes}  & No   \\ \cline{2-4} 
                                  & \multicolumn{1}{c|}{Yes} & \multicolumn{1}{c|}{18.4\%} & 16.8\% \\ \cline{2-4} 
                                  & \multicolumn{1}{c|}{No}  & \multicolumn{1}{c|}{32.7\%} & 32.0\% \\ \hline
    \end{tabular}
    \label{tab:mk-agreement}
    \caption{Monster Killer}
\end{subtable}

\caption{The agreement and disagreement between Self-Perceived labels and the AAR labels for every persona.}
\label{tab:aar-sl-agreement}
\vspace{-10pt}
\end{table*}

Table~\ref{tab:aar-label} displays the breakdown of player distributions using every AAR multilabel combination. One interesting finding is that there is a sizable group of playtraces ($82$) with no label at all, suggesting the presence of a fourth persona not measured here. Another curious group of players are the $117$ traces that are classified as all three persona types. The mechanics showcased here are selected due to their traditional correlation with a specific persona, and values in \textbf{bold} correspond to that combination's correlated mechanics.

All combinations which include the ``R'' label contain a lower amount of steps taken than other label combinations, the lowest coming from ``Pure R'' at $18.49$ and ``R\&MK'' at $15.25$. ``Pure TC'' and ``Pure MK'' distributions contain similar amounts of treasures collected and monsters killed, but there are only $2$ ``Pure MK'' playtraces in the entire dataset. The presence of many ``R\&MK'' ($56$), ``TC\&MK'' ($24$) and ``R\&TC\&MK'' playtraces but few ``Pure MK'' traces promotes the idea that players that are driven to kill monsters are also driven toward other persona subgoals during gameplay. The highest combinations of treasures collected and enemies killed comes from the ``TC\&MK'' labeled traces, $6.83$ and $5.0$ respectively, suggesting TC and MK subgoals have great synergy. The ``R\&TC'' traces have a greater amount of enemies killed than ``R\&MK,'' but the treasure collection amounts differ greatly ($3.36$ and $0.18$ respectively). Due to the nature of enemy movement in Minidungeons 2 and the placement of enemies and treasure on some of the user study maps (Section \ref{sec:experiment-maps}), it is difficult to collect treasure and not kill monsters. However, it is quite simple to kill monsters and \textit{not} collect the nearby treasure, which is immobile. We think that ``R\&TC'' players will kill monsters on the way to collect treasure, whereas ``R\&MK'' players kill monsters and ignore the treasure. Also of note is the different in steps taken between the two distributions: ``R\&TC'' players ($36.82$) take over twice as many steps as ``R\&MK'' players ($15.25$) in order to collect more treasure. This result suggests that ``R\&TC'' players may have competing subgoals: ``go to the exit as fast as possible, but also go out of your way to collect treasure.''

Table~\ref{tab:self-label} displays the breakdown of player distributions using the labels coming from the questionnaire answers for all multilabel combinations. We notice that more than 60\% of all the playtraces are either `No Label' or `R \& TC \& MK'. Looking at the different distributions of `Steps Taken', `Treasure Collected', and `Enemies Killed', we noticed that any combination with runner persona have the lowest number of `Steps Taken' compared to all the other similar to AAR labels. We think that runner players have a high tendency to not change their persona compared to others. On the other hand, we think other personas change themselves to become a runner while playing the game. This can be seen by comparing the number of playtraces that are classified as runner by AAR label compared to the self perceived labels. Another noticable thing is the distributions for all the personas have a higher standard deviations compared to the AAR labels which confirms our point that users are either changing their personas during playing or can't identify themselves correctly. Finally, both `Treasure Collected' and `Enemies Killed' averages are way lower than the ones shown shown by the AAR labels. This suggests that users who think themselves as a monster killer or a treasure collector actually behave more like a Runner, as the mean of `Steps Taken' mechanic is low by comparison.

The questionnaire data presents a rare opportunity to compare human self-perception with a synthetically generated ground truth. To gauge the accuracy of self-perception, we compare the self-perceived label with the multilabel generated from action agreement. We calculate the amount of times that both labels either said a playtrace was a persona or both said a playtrace was not a persona. Self-perceived and action agreement labels both agree (both are ``Yes'' or ``No'' at the same time) on Runner labels $47.3\%$ of the time, Treasure Collectors $59.1\%$ of the time, and Monster Killers only $50.4\%$ of the time. For all three persona styles, agreement is a coin toss. The confusion matrices (Table \ref{tab:aar-sl-agreement}) between AAR and Self-Perceived labels demonstrates this disagreement across each of the persona types.

The discrepancy between perception and action agreement labels suggests a possible cognitive dissonance between how players believe that they play and how they actually do play. Another possible reason for this disagreement is that users tend to have dynamic goals, unlike static personas. Players can change their play persona to adapt to any factor be it environmental rewards (the level might be easier or more enjoyable for a treasure collector persona over another) or agential motivations (the player is bored and decided to race through the level like a runner). Lastly, the participants of this study may be more familiar with other games in which they behave how they perceive themselves to behave. Since Minidungeons 2 is a new environment for them, their lack of familiarity may have caused them to behave differently.

\subsection{Classifier Training and Testing}

Table \ref{tab:training-details} displays the results of training and validation of the classifiers on the synthetic and human datasets. The SVM trained on synthetic data clearly fail to capture the distribution of the human players. Although there are $1,500$ playtraces to train on with different distribution, the synthetic playtraces distribution is far away from human data distribution. We think due to the purity of the synthetic agents (only runner, only monster killer, or only treasure collector), the trained classifier fails to recognize any multi-label persona. As a result, they do not perform well on the human testing set with an accuracy of $4.8\%$. The LSTM model is also not able to generalize well on the synthetic data, having a low accuracy during training ($58.1\%$) and testing ($18.6\%$).
% The LSTM model is not capturing the essence of a player with multiple labels and ends up predicting the entire test set to be one class at a time. 
% This also supports the hypothesis that the procedural personas do not behave like humans do, at least on a global level. Rarely does a human hyperfocus on a specific goal during gameplay and ignore other subgoals like a procedural persona might.
During training, the LSTM is not able to differentiate between treasure collector and monster killer personas (due to proximity of treasures and monsters) which causes it to classify all the treasure collector as monster killers during training. We believe that the lack of variety between playtraces explains this poor performance for both the LSTM and the SVM.

\begin{table}[ht]
\centering
\resizebox{\columnwidth}{!}{%
\begin{tabular}{|c|c|l|l|l|}
\hline
\textit{Model} & \textit{Training Set} & \multicolumn{1}{c|}{\textit{Training}} & \multicolumn{1}{c|}{\textit{Validation}} & \multicolumn{1}{c|}{\textit{Testing}} \\ \hline
\hline
\multirow{3}{*}{LSTM} & \textit{Synthetic} & 0.581 ± 0.047 & 0.483 ± 0.08 & 0.186 ± 0.029 \\ 
\cline{2-5}
& \textit{Human SL}  & 0.563 ± 0.016 & 0.48 ± 0.112 & 0.187 ± 0.014\\
\cline{2-5}
& \textit{Human AAR}  & 0.837 ± 0.03 & 0.784 ± 0.067 & \textbf{0.726 ± 0.029}    \\
\hline
\hline
\multirow{3}{*}{SVM} & \textit{Synthetic} & 0.596      & 0.567        & 0.048        \\
\cline{2-5}
& \textit{Human SL}  & 0.43 & 0.359      & 0.259        \\
\cline{2-5}
& \textit{Human AAR}  & 0.777         & 0.694      & \textbf{0.700}    \\
\hline
\end{tabular}%
}
\caption{The training, validation, testing results for the LSTM and SVM trained on different datasets/labels}
\label{tab:training-details}
\vspace{-10pt}
\end{table}

Both the SVM and the LSTM models trained on the human dataset labeled using self-perception (SL) struggles to find a pattern in the training data and also performs poorly on the testing set with an accuracy of $25.9\%$ and $18.7\%$ respectively. This result supports the finding in Section \ref{sec:label-comparison}, that humans may be poor judges of their own play persona.

However, the SVM and LSTM models trained on the human dataset labeled using AAR are able to generalize to perform well on the test set, with an accuracy of $70\%$ and $72.6\%$ respectively. The testing accuracy is also similar to its training/validation scores, meaning that both distributions of human players, which total $363$ people, contain similar mechanical patterns that the classifier picked up on. The similarity in scores between LSTM model and SVM model supports our hypothesis that mechanic frequency is a good data augmentation method, being able to capture the essence of the play persona similar to using playtrace data.

\section{Conclusion \& Future Work}
In this paper, we showcased that training a machine learning model on user data that is labeled using action agreement ratio is capable to generalize and predict the play personas of unseen players.
We use two different input representations: a cropped sequence of states and mechanic frequencies, showcasing that both can be used interchangeably depending on what is easier to capture in the game. We demonstrate that just using synthetic data from the agents or self-labeled playtraces fails to generalize well. By modeling with a small human AAR labeled dataset, we can successfully predict the personas of unseen human playtraces. Game states and mechanic frequencies are problem independent, meaning that all they require is the user to define what a \textit{state} or a \textit{mechanic} is as well as build a classifier that can take them as input. These are usually defined during the game design process anyways and typically do not require more work from developers.

Calculating the actual labels using AAR took on average $40$ seconds per playtrace with parallelization and almost 7 hours for the $8$ trace training set in entirety. For comparison, training the SVM on mechanic frequency vectors took less than $5$ seconds overall and can infer on new data in less than a second. Time is not usually a concern during development. But in published games with millions of active players playing daily it becomes infeasible to calculate play personas using AAR. At the speed at which we labeled the $853$ trace dataset using AAR, labeling of a $10,000$ trace dataset takes $4.5$ days. 
% If AAR integrated into a labeling pipeline for a game like Candy Crush, at this speed we would be ingesting less than 0.01\% of daily active players\footnote{Based on a daily active player average of 260,000,000, taken from https://activeplayer.io/candy-crush-saga/
% on 05/22/2022}.

% Companies could always collect data from the testing teams and then label it using AAR then train these models to be an assistant tools to designers and developers.

The importance of identifying persona goes beyond just testing if the levels have the required experience but also generating personalized levels~\cite{green2020mario} and/or tutorials~\cite{green2018atdelfi}.
% Having a fast classification engine would help generative systems by speeding up the generation and make sure the development cycle goes fast. We think this avenue is a great direction for our future work. 
Outside of PCG, play persona statistics can be a valuable source of feedback for designers to improve and iterate levels. Play personas can easily be integrated in the development engines as tool to direct designers about map and level experience.

\section*{Acknowledgment}
We would like to thank those that took part in our user study and took the time to answer the questionnaire.

\bibliographystyle{IEEEtran}
\bibliography{references}
\end{document}